\documentclass[10pt,twocolumn]{article}
\usepackage{pdfpages}

\usepackage{lmodern, upquote}
\usepackage{geometry}
\usepackage{titlesec}
\usepackage{makecell}

\titlespacing{\section}{0em}{*1}{*1}[0em]
\titlespacing{\subsection}{0em}{*1}{*0.7}[0em]
\titlespacing{\subsubsection}{0em}{*0.8}{*0.5}[0em]
\titlespacing{\paragraph}{0em}{*0.5}{*0.5}[0em]

\renewenvironment{abstract}{\subsection*{Abstract}\em}{}
\newenvironment{keywords}{\subsection*{Keywords}\em}{}

\usepackage{amsmath, amsfonts, amssymb}
\usepackage{graphicx}
\graphicspath{{figures/}}
\usepackage[hyphens]{url}
\usepackage{hyperref}
\hypersetup{colorlinks=true}
\usepackage{multirow, siunitx}
\usepackage{placeins} 
\usepackage{cleveref}

\title{Towards a Data-Driven Requirements Engineering Approach:\\ Automatic Analysis of User Reviews}

\author{Jialiang Wei$^*$, Anne-Lise Courbis$^*$, Thomas Lambolais$^*$, Binbin Xu$^*$, \\Pierre Louis Bernard$^{**}$ and Gérard Dray$^*$
\\[6pt]
$^{*}$: {\normalsize EuroMov Digital Health in Motion, Univ Montpellier, IMT Mines Ales, Ales, France}\\
$^{**}$: {\normalsize EuroMov Digital Health in Motion, Univ Montpellier, IMT Mines Ales, Montpellier, France}\\
\texttt{{\small jialiang.wei@mines-ales.fr}}
}
\date{}

\begin{document}

\maketitle

\textbf{This is the translated English version, the original work was written in French, attached in this PDF:  {\hypersetup{linkcolor=blue}\hyperlink{page.5}{\emph{Go to page 5}}}}

\begin{abstract}
We are concerned by Data Driven Requirements Engineering, and in particular the consideration of user's reviews. These online reviews are a rich source of information for extracting new needs and improvement requests. In this work, we provide an automated analysis using CamemBERT, which is a state-of-the-art language model in French. We created a multi-label classification dataset of 6000 user reviews from three applications in the Health \& Fitness field\footnote{Dataset is available at: \url{https://github.com/Jl-wei/APIA2022-French-user-reviews-classification-dataset}}. The results are encouraging and suggest that it's possible to identify automatically the reviews concerning requests for new features.
\end{abstract}

\footnotetext[2]{7ème Conférence Nationale sur les Applications Pratiques de l’Intelligence Artificielle}


\section{Introduction}

Requirements Engineering (RE) aims to ensure that systems meet the needs of their stakeholders including users, sponsors, and customers. Placed in the preliminary and upstream stage of Software Engineering cycle, it plays a vital role in ensuring the software quality \cite{Lamsweerde2009}. RE includes requirements elicitation, requirements modeling \& analysis, requirements verification and requirements management. Requirements elicitation is the first step of requirements engineering, which aims to discover the real needs of stakeholders \cite{Bennaceur2019}. Conventional approaches aimed at establishing explicit models of requirements based on interviews, brainstorming, observations, etc. are now enriched by new ones which directly exploit the user feedback posted on app markets or social networks. Thus, the RE community proposes to take up new challenges \cite{Franch2021} consisting in developing a Data Driven Requirements Engineering allowing to process a large volume of user feedback.

The app markets, like Google Play and App Store, provide a valuable channel for users and app developers to communicate, through which users express their comments on apps -- including praise and criticism of the application, user experience, bug report and feature requests \cite{6636712}. Many of these feedback are helpful and can serve for requirements elicitation. For example, Facebook app receives more than 4000 user reviews per day, of which 30\% can be used for requirements elicitation \cite{6636712}. While manually sifting through these feedback is laborious. To filter out the most useful user feedback, we employ Bidirectional Encoder Representations from Transformers (BERT) \cite{Devlin2019}, a deep learning model based on Transformer architecture. The pre-trained BERT model can be fine-tuned for different downstream tasks, including text classification, question answering, natural language inference, etc. Models based on BERT have achieved the state-of-the-art metric on almost all natural language processing (NLP) tasks. The original BERT is pre-trained with English corpus, while CamenBERT \cite{Martin2020} is a language model for French which is pre-trained on the French sub-corpus of the OSCAR. 

For BERT model fine-tuning, new applications should be trained/retrained with labelled data from downstream tasks. There exists several user reviews dataset for classification task \cite{8048885,8057860,7372065,Palomba2018,7884612,Maalej2016, 8933719}. They are all in English, two of them are multilingual \cite{8057860, 8933719}, but none of them contains enough reviews in French. 
To fill this gap, we created a French user reviews dataset including 6000 reviews for multi-label classification task, it contains four labels: rating, bug report, feature request and user experience. As the main theme of our project is data driven development applied in system monitoring on the elderly for 

\section{Method}
\label{sec:methoddata}

\subsection{Dataset}

NLP tasks based on CamemBERT model require large number of labelled user reviews on apps in French. To the best of our knowledge, such dataset does not exist yet. This drives us to create French user reviews dataset and to share it with the community. Firstly, we collected the French user reviews on three applications from Google Play: Garmin Connect, Huawei Health and Samsung Health. The number of reviews collected and the sampled size to constitute the \textit{dataset} during the learning phase are presented in Table~\ref{tab:dataset}. Then we manually labelled all the sampled user reviews. As proposed in the work of~\cite{Maalej2016}, we selected four labels: \textit{rating}, \textit{bug report}, \textit{feature request} and \textit{user experience}. \textit{Ratings} are simple text which express the overall evaluation to that app, including praise, criticism, or dissuasion. \textit{Bug reports} show the problems that users have met while using the app, like data loss, app crash, connection error, etc. \textit{Feature requests} category reflects the demand of users on new functionality, new content, new interface, etc. In \textit{user experience}, users describe their practical experience in relation to the functionality of the app, how certain functions are helpful. As we can observe from Table~\ref{tab:reviewExamples}, it shows examples of labelled user reviews, each review can belong to one or more categories resulting a multi-label classification problem. 

\begin{table}[!htb]
\centering
 \begin{tabular}{c c c c}
 \hline
 App & \#Reviews & Sample \\
 \hline
 Garmin Connect & 22880 & 2000 \\ 
 Huawei Health & 10304 & 2000 \\
 Samsung Health & 18400 & 2000 \\
 \hline
 \end{tabular}
 \caption{Applications and collected reviews for the pre-training phase}
\label{tab:dataset}
\end{table}

\begin{table}[!htb]
\centering
 \begin{tabular}{c c c c c c} 
 \hline
 App & Total & (R) & (B) & (F) & (U) \\
 \hline
 Garmin Connect & 2000 & 1260 & 757 & 170 & 493 \\ 
 Huawei Health & 2000 & 1068 & 819 & 384 & 289 \\
 Samsung Health & 2000 & 1324 & 491 & 486 & 349 \\
 \hline
 \end{tabular}
 \caption{Manual classification of reviews into 4 families: (R)ating, (B)ug report, (F)eatures request, (U)ser experience }
\label{tab:dataLabel}
\end{table}

\begin{table*}[!htb]
\renewcommand{\arraystretch}{1.3}
\centering
 \begin{tabular}{p{.7\textwidth} | c} 
 \hline
 \multicolumn{1}{c|}{\textbf{User review}} & \textbf{Labels} \\
 \hline
 \texttt{Très bien} & Rating \\
 \hline
 \texttt{très bon application qui aide à faire plus activités} & Rating, User Experience \\
 \hline
 \texttt{C'est une très bonne application. Elle nous aide  beaucoup à rester toujours actif.} & Rating, User Experience \\
 \hline
 \texttt{Cette appli prends bien en compte les trajets et son système de défi permet de bien se motiver à marcher.} & User Experience \\
 \hline
 \texttt{j'aimais bien cette appli mais elle ne fonctionne plus: message d'erreur téléphone rooté !! faux et j'ai vérifié.} & Rating, Bug Report \\
 \hline
 \texttt{Hello. It is possible to add the STOP SMOKING function in your application with a Widget for the S3 Frontier watch. Thank you} & Feature Request \\
 \hline
 \end{tabular}
 \caption{Examples of labelled user reviews}
\label{tab:reviewExamples}
\end{table*}

The user reviews are annotated by four authors of this paper and then reviewed by Jialiang Wei.
The number of reviews classified in each of the categories for each of the applications is given in Table~\ref{tab:dataLabel}. The sum of classified reviews does not equal the total of reviews because some reviews have been assigned to more than one class.

\subsection{Model}
The user reviews classification is a common text classification problem in NLP, which aims to assign labels to textual units \cite{10.1145/3439726}. Due to the large amount of text data, manual classification is unpractical, automatic classification is becoming more and more important or necessary in real applications. 
In recent years, user reviews classification problem has been addressed with many machine learning (ML) or deep learning (DL) approaches. 
Maalej et al. \cite{Maalej2016} used Naïve Bayes, Decision Tree and Maximum Entropy to classify user reviews into bug reports, feature requests, user experiences, and text ratings. 
Restrepo Henao et al. \cite{9582367} applied BERT to classify user reviews in three categories(problem report, feature request and irrelevant), and they compared the performance of conventional ML, CNN-based DL model, BERT model and found that BERT allows obtaining the highest precision and recall. Mekala et al. \cite{9604705} compared the performance of crowdsourcing, SVM, Naïve Bayes, FastText-based classifier, ELMo and BERT on classifying useless and helpful app reviews. Among all these methods, BERT performed the best overall.
As BERT archived the best result on classification of user feedback, we decided to utilize BERT for the classification task in this work. And we use CamemBERT \cite{Martin2020}, model trained with a large French corpus, for classifying our French user reviews dataset. 

\begin{figure}[!htb]
\centerline{
    \includegraphics[width=\columnwidth]{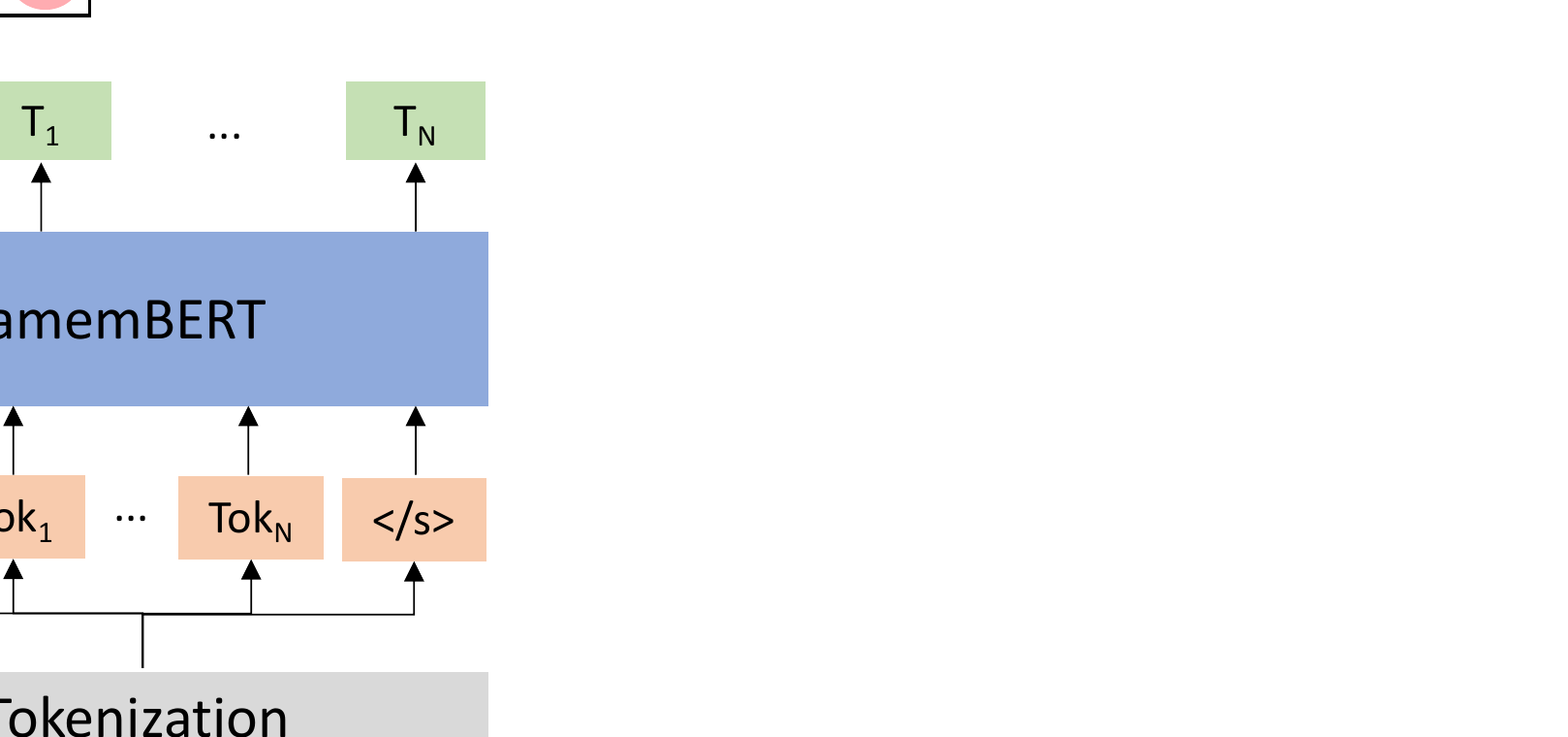}}
    \caption{CamemBERT for multi-label classification}
    \label{fig:camembert}
\end{figure}

CamemBERT cannot use raw text as input, each user review is at first tokenized into subword units \cite{sennrich-etal-2016-neural}, and '\textless s\textgreater' and '\textless /s\textgreater' is added at the start and end of the review. The tokenized review is then padded to a length of 512 with a special ‘\textless PAD\textgreater’ token, and attention masks is added for each of ‘\textless PAD\textgreater’ token. Finally, each subword token is mapped to a numeric ID. To convert the CamemBERT representation to the classification task, we have also adapted the architecture of CamemBERT by adding a linear layer with 4 outputs on the its top, as shown in figure \ref{fig:camembert}

\section{Experiment}
\label{sec:experiment}
For our experiments, we used pre-trained CamemBERT model from HuggingFace platform. Each experience was trained for 3 epochs, with a batch size of 1 and used the AdamW optimizer with a learning rate $2e^{-5}$ on a machine with 48 GB RAM, an Intel i7 processor and NVIDIA Quadro M2200 GPU with 4 GB VRAM. To evaluate the performance of CamemBERT on user feedback classification, we designed two types of experiments. In the first experiments, we trained and tested the model with the user reviews of the three apps. While in the second type, we trained the model with the user reviews of two apps and compared the performance of the model on the user reviews of these two apps and on the user reviews of the third app.

\subsection{Train and test on user reviews from same apps}
In this experiment, we split 60\% of the 6000 user reviews as training set, 20\% as validation set and 20\% as test set in a stratified manner. For the purpose of cross validation, each model is trained 10 times with random split (stratified sampling) of training and test sets. As for the evaluation metric, we use precision, recall and F1 score. The mean performance of 10 runs is shown in Table~\ref{tab:result1}.

\begin{table}[!htb]
\centering
 \begin{tabular}{l c c c} 
 \hline
 & Precision & Recall & F1 \\
 \hline
Rating & 0.88 & 0.93 & 0.91 \\
Bug Report & 0.92 & 0.93 & 0.93 \\
Feature Request & 0.85 & 0.83 & 0.84 \\
User Experience & 0.81 & 0.73 & 0.77 \\
\textbf{Weighted Average} & 0.88 & 0.89 & 0.89 \\
 \hline
 \end{tabular}
 \caption{Classification accuracy on user reviews of three apps}
\label{tab:result1}
\end{table}

As shown in Table~\ref{tab:result2}, the retrained model archives good results overall, with a weighted average F1 0.89. While the performance on \textit{user experience} is relatively lower than the other three categories. According to our labeling experiences, there would be two reasons. Firstly, the \textit{user experience} are more diverse than three other categories, the model may need a larger training set to learn the features of that type. The second reason could probably lie in the fact that four authors participated in the annotation of the reviews and their interpretation of the \textit{user experience} criterion can influence the performance of the model, even though the same definition / rule is applied. 

\subsection{Train and test on user reviews from separate apps}
In this experiment, we used the same stratified sampling strategy: 60\% of the reviews of two apps as training set, 20\% as validation. The tests apply to the remaining 20\% and also to all reviews of the third application. 
For example, one of these combinations consisted in running the training / validation on 1600 reviews of Garmin Connect and 1600 of Huawei Health and the test on the 800 reviews of these applications and the 2000 reviews of Samsung Health.
There are three combination of apps, for each combination, we ran 10 times. The mean results of the 30 runs ($3 \times 10$) is shown in Table~\ref{tab:result2} and Table~\ref{tab:result3}.

\begin{table}[!htb]
\centering
 \begin{tabular}{l c c c} 
 \hline
 & Precision & Recall & F1 \\
 \hline
Rating & 0.88 & 0.93 & 0.90 \\
Bug Report & 0.91 & 0.93 & 0.92 \\
Feature Request & 0.85 & 0.81 & 0.83 \\
User Experience & 0.79 & 0.72 & 0.76 \\
\textbf{Weighted Average} & 0.87 & 0.88 & 0.88 \\
 \hline
 \end{tabular}
 \caption{Classification accuracy on 20\% rest user reviews of two apps}
\label{tab:result2}
\end{table}

\begin{table}[!htb]
\centering
 \begin{tabular}{l c c c} 
 \hline
 & Precision & Recall & F1 \\
 \hline
Rating & 0.88 & 0.92 & 0.90 \\
Bug Report & 0.85 & 0.92 & 0.88 \\
Feature Request & 0.80 & 0.74 & 0.75 \\
User Experience & 0.77 & 0.69 & 0.73 \\
\textbf{Weighted Average} & 0.86 & 0.86 & 0.85 \\
 \hline
 \end{tabular}
 \caption{Classification accuracy on all user reviews of third apps}
\label{tab:result3}
\end{table}

Performances in Table~\ref{tab:result2} are slightly lower to those in Table~\ref{tab:result1}, which means that we can get good accuracy with even smaller training set. As shown in Table~\ref{tab:result3}, the average precision and recall on unseen apps' reviews is just $0.01-0.02$ ($1\%$ to $2\%$) lower than on the apps which we have trained on.

\section{Discussion \& Future Work}
In this work, we created a dataset for the multi-label classification of reviews written in French by users of physical activity monitoring applications. We used the CamemBERT model to classify these reviews and the results showed good performance which confirmed the experience in other works. The training and test experiments on different applications showed the feasibility of generalizing the model on applications from the same App category. These results encourage us to continue our work on Data Driven Requirements Engineering for the development of different versions of an application for preventing the adverse aging effects by monitoring the physical activity of seniors in a context of \emph{healthy aging}. We plan to refine the classification of requests for new functionalities by making it possible to identify, through an unsupervised approach, the concepts of the domain with respect to the target application, such as walking speeds and times, imbalances, sleep monitoring, etc. This classification could provide a visual scheme of the functionalities requested by the users for App designer. 

The overall objectives of the project will be to (i) group the requests and proposals by category of requirements, (ii) match the identified requirements with the models present in App specifications, in order to be able to take them all into account during the design phase with a more relevant and optimised manner.

\Urlmuskip=0mu plus 1mu\relax
\renewcommand\refname{References}
\bibliographystyle{ieeetransnourl}
\small{\bibliography{refsv3}}

\clearpage



\section*{Supplementary material}

# Vers une ingénierie des exigences dirigée par les données : analyse automatique d'avis d'utilisateurs

Jialiang Wei*, Anne-Lise Courbis*, Thomas Lambolais*, Binbin Xu*,
Pierre Louis Bernard** and Gérard Dray*

*: EuroMov Digital Health in Motion, Univ Montpellier, IMT Mines Ales, Ales, France
**: EuroMov Digital Health in Motion, Univ Montpellier, IMT Mines Ales, Montpellier, France
*: firstname.lastname@mines-ales.fr **: firstname.lastname@umontpellier.fr

**Résumé**

*Ces travaux s'inscrivent dans le cadre de l'ingénierie des exigences dirigée par les données et notamment les avis d'utilisation. Les commentaires en ligne d'utilisateurs d'applications sont une source importante d'informations pour en améliorer leur fonctionnement et en extraire de nouveaux besoins. Nous proposons une analyse automatisée en utilisant CamemBERT, un modèle de langue en français qui fait aujourd'hui état de l'art. Afin d'affiner ce modèle, nous avons créé un jeu de données de classification multi-labels de 6000 commentaires issus de trois applications du domaine de l'activité physique et la santé [1]. Les résultats sont encourageants et permettent d'identifier les avis concernant des demandes de nouvelles fonctionnalités.*

**Mots-clés**

*Ingéniérie des exigences, Ingénierie guidée par les données, TALN, CamemBERT, Apprentissage profond*

**Abstract**

*We are concerned by Data Driven Requirements Engineering, and in particular the consideration of user's reviews. These online reviews are an important source of information for extracting new needs and improvement requests. In this paper, we provide an automated analysis using CamemBERT, which is a state-of-the-art language model in French. In order to fine tune the model, we created a multi-label classification dataset of 6000 user reviews from three applications in the Health & Fitness field. The results are encouraging and make it possible to identify automatically the reviews concerning requests for new features.*

**Keywords**

*Requirements Engineering, Data Driven Requirements Engineering, NLP, CamemBERT, Deep Learning*

---

1. Données disponible en ligne : https://github.com/Jl-wei/APIA2022-French-user-reviews-classification-dataset

# 1 Introduction

L'ingénierie des exigences (IE) est l'une des phases du cycle de développement des systèmes, visant à élaborer un cahier des charges aussi clair et complet que possible, à partir des interviews de différentes parties prenantes (clients, utilisateurs, donneurs d'ordre). Il s'agit d'une des phases les plus critiques [1], jouant un rôle fondamental pour obtenir un logiciel de qualité. Les activités principales de l'IE sont : la conduite des interviews, l'élicitation, la modélisation du domaine et l'analyse de faisabilité. L'étape d'élicitation a pour objectif de mettre en exergue les besoins réels des parties prenantes [2]. Les approches traditionnelles visant à établir des modèles explicites d'exigences à partir d'interviews, brainstorming, observations, etc. sont enrichies par de nouvelles approches, visant à exploiter les retours des utilisateurs mis en ligne sur les stores d'applications ou les réseaux sociaux. Ainsi, la communauté d'IE propose de relever de nouveaux défis [3] consistant à développer une ingénierie des exigences dirigée par les données permettant de traiter un volume important d'avis d'utilisateurs.

Les stores d'applications, comme Google Play® et App Store®, sont devenus des plate-formes de communication entre utilisateurs et développeurs assurant la collecte d'avis sur chacune des applications en téléchargement. Ces avis sont riches en information car ils contiennent des critiques et des préférences, des retours d'expérience, des rapports d'erreur et des demandes de nouvelles fonctionnalités. Par exemple, l'application Facebook reçoit plus de 4000 commentaires journaliers dont 30% peuvent être considérés comme une source pour identifier de nouveaux besoins [4]. Exploiter manuellement de telles sources serait laborieux. Nous proposons d'effectuer un Traitement Automatique du Langage Naturel (TALN) pour filtrer les avis et cibler les plus pertinents en vue d'identifier de nouvelles exigences. Plus précisément, nous nous sommes intéressés à BERT (*Bidirectional Encoder Representations from Transformers*) [5], qui est un modèle d'apprentissage profond basé sur une architecture de type *Transformer*. BERT peut être affiné pour effectuer différentes tâches telles que la classification de textes, la réponse aux questions, l'inférence en langage naturel, etc. Ce sont les modèles basés

Vers une ingénierie des exigences dirigée par les données : analyse automatique d'avis d'utilisateurs

sur BERT qui ont obtenu les meilleurs résultats sur la plupart des tâches de la communauté TALN. BERT étant entraîné en anglais, nous nous sommes tournés vers CamemBERT [6], un modèle de type BERT qui a été entraîné sur le corpus OSCAR en français afin d'avoir de meilleures performances sur les commentaires écrits en français.

Pour affiner le modèle BERT, il est nécessaire de l'entraîner avec un ensemble de données annoté. Il existe plusieurs jeux de données de commentaires annotés pour effectuer la classification [7–13]. La plupart de ces jeux de données sont en anglais. Certains sont multi-langues [8, 13]. Mais aucun d'eux n'intègre suffisamment de commentaires rédigés en français. Nous proposons donc de créer un jeu de données en français et d'explorer les résultats obtenus. Notre sujet d'étude étant à terme le développement selon une approche dirigée par les données d'applications de suivi de séniors pour « *bien vieillir en santé* », nous nous sommes intéressés aux commentaires de trois applications de la catégorie « Health & Fitness » de Google Play.

Dans la partie suivante nous présentons le jeu de données constitué ainsi que la méthode mise en œuvre pour conduire notre expérimentation. Les résultats de classification obtenus sont discutés dans la troisième partie. Enfin, dans la dernière partie, nous exposons les perspectives envisagées sur ces travaux.

## 2 Méthode mise en œuvre pour l'apprentissage

### 2.1 Jeu de données et son annotation

L'usage de CamemBERT pour classifier les avis d'utilisateurs écrits en français nécessite un jeu de données annoté. En l'état actuel de nos connaissances, il n'en existe pas. Nous en avons donc créé un. Nous avons tout d'abord collecté sur la plateforme Google Play des avis rédigés en français concernant les applications Garmin Connect, Huawei Health et Samsung Health. Le nombre de commentaires collectés et la taille choisie pour constituer le jeu de données pour la phase d'apprentissage sont présentés dans le Tableau 1. Nous avons ensuite associé manuellement des labels aux avis. Nous avons choisi d'utiliser quatre labels, comme cela est proposé dans [12] : *Évaluation*, *Rapport d'erreur*, *Demande de nouvelles fonctionnalités* et *Expérience utilisateur*. L'*évaluation* est généralement un texte simple qui exprime le sentiment général de l'utilisateur sous forme d'éloge, de critique ou de dissuasion. Le *rapport d'erreur* est relatif aux problèmes rencontrés lors de l'utilisation de l'application : perte de données, arrêt brutal de l'application, problème de connexion, etc. La *demande de nouvelles fonctionnalités* concerne de nouveaux services, de nouveaux contenus ou encore de nouvelles interfaces. Enfin, les avis de type *expérience utilisateur* sont des expériences relatées par les utilisateurs qui sont décrites par exemple comme des conseils d'utilisation, des fonctions ou des usages perçus comme très utiles ou faciles d'utilisation. Il est possible d'associer plusieurs labels à un même avis. Par exemple, l'avis : *"c'est une très bonne application. Elle nous aide beaucoup à rester toujours actif."* sera annoté dans les catégories *Évaluation* et *Expérience Utilisateur*. L'avis *"j'aimais bien cette application mais elle ne fonctionne plus : message d'erreur téléphone rooté !! faux et j'ai vérifié."* sera annoté dans les catégories *Évaluation* et *Rapport d'erreur*. Les avis sont annotés par quatre auteurs de l'article, et révisés par Jialiang Wei. Le Tableau 2 indique le nombre d'avis classés dans chacune des catégories pour les trois applications cibles. La somme des avis classés n'est pas égale au total des avis puisque certains avis ont été affectés à plusieurs classes.

| App | #Avis | Échantillon |
|---|---|---|
| Garmin Connect | 22880 | 2000 |
| Huawei Health | 10304 | 2000 |
| Samsung Health | 18400 | 2000 |

TABLEAU 1 – Applications et avis collectés pour la phase d'entraînement

| App | Total | (Ev) | (R) | (D) | (Exp) |
|---|---|---|---|---|---|
| Garmin Connect | 2000 | 1260 | 757 | 170 | 493 |
| Huawei Health | 2000 | 1068 | 819 | 384 | 289 |
| Samsung Health | 2000 | 1324 | 491 | 486 | 349 |

TABLEAU 2 – Classification manuelle des avis en quatre familles : (**Ev**)aluation, (**R**)apport d'erreur, (**D**)emande de fonctions, (**Exp**)érience utilisateur

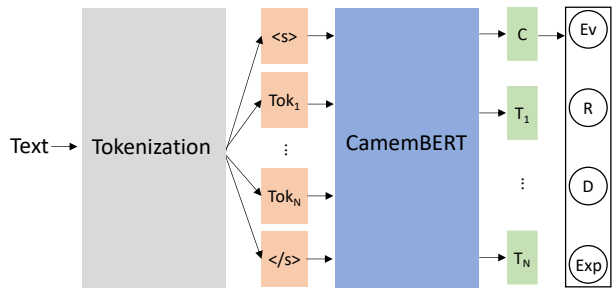

FIGURE 1 – Vue d'ensemble du modèle CamemBERT pour la classification multilabels

### 2.2 Modèle de classification

La classification des avis d'utilisateurs est un problème de classification de type TALN [14]. Ces dernières années, de nombreuses méthodes d'apprentissage statistique ont été appliquées à la classification d'avis. Maalej et al. [12] utilise des réseaux bayésiens, des arbres de décision et régression logistique multinomiale pour effectuer la classification des avis dans ces mêmes quatre catégories : *évaluation*, *rapport d'erreur*, *demande de fonctionnalités* et *expérience utilisateur*. Restrepo Henao et al. [15] applique BERT pour faire la classification des avis dans les catégories : *rapport de problèmes*, *demande de fonctionnalités*, et *avis non pertinent* ; ils comparent les performances des modèles d'apprentissage BERT et d'un réseau de neurones convolutif à apprentissage profond et concluent que BERT obtient les

APIA@PFIA 2022   120

meilleures taux de précision et de rappel. Mekala et al. [16] comparent la performance dans le domaine du *crowdsourcing* (production participative) de BERT avec des modèles de Machines à Vecteurs de Support, des classifieurs de type FastText et ELMo sur des avis d'utilisateurs classés en deux catégories : *inutile* et *utile*. BERT obtient également les meilleurs résultats. Les bonnes performances de BERT ont guidé notre choix sur sa version française CamemBERT. CamemBERT ne peut utiliser du texte brut en entrée. Une phase de pré-traitement est nécessaire : chaque avis est découpé en un ensemble de termes, ou jetons [17]. Les balises `<s>` et `</s>` sont ajoutées pour marquer respectivement le début et la fin de l'avis. Les avis sont ensuite formatés à la longueur 512 en complétant ceux qui sont de taille inférieure avec la balise `<PAD>`. Ceux qui sont de taille supérieure sont tronqués. Un masque est alors associé à chaque avis afin de repérer les symboles `<PAD>` (valeur 0) et les termes (valeur 1). Puis, on attribue à chaque terme un identifiant numérique. Enfin, pour adapter CamemBERT à la tâche de classification souhaitée, nous modifions son architecture en ajoutant à son dernier niveau une couche linéaire à quatre sorties, comme le montre la Figure 1.

## 3 Expérimentations et résultats

Nous avons utilisé la bibliothèque PyTorch pour la mise en œuvre CamemBERT. Le modèle a été entraîné avec trois epochs, le paramètre `batch_size` et l'optimiseur AdamW avec un taux d'apprentissage fixé à $2e^{-5}$. Le modèle a été entraîné sur un ordinateur de 48 Go de mémoire RAM, muni d'un processeur Intel i7-7820HQ et d'une carte graphique NVIDIA Quadro M2200 de 4 Go de VRAM. L'évaluation de la performance de CamemBERT sur la classification des avis s'est faite sur deux expérimentations. Dans la première, les modèles sont entraînés sur une partie des avis des trois applications et testés sur l'autre partie. Dans la seconde expérimentation, les modèles sont entraînés sur les avis de deux applications et évalués sur ces applications et sur la troisième application.

### 3.1 Apprentissage à partir des trois applications

Pour cette expérimentation, nous avons sélectionné 60% des avis des trois applications comme ensemble d'apprentissage, 20% comme ensemble de validation et 20% comme ensemble de test en utilisant un échantillonnage stratifié. Cette opération a été répétée 10 fois. Les critères d'évaluation de performance des modèles sont : la précision, le rappel et la F1-Mesure. Les résultats moyens obtenus sur les 10 apprentissages sont présentés dans le Tableau 3.

Comme le montre le Tableau 3, les résultats obtenus sont corrects dans l'ensemble : la moyenne de la F1-Mesure est de 0,89. Les résultats concernant le critère *Expérience Utilisateur* sont inférieurs aux trois autres. Ceci s'explique de deux façons. Tout d'abord les avis concernant ce critère sont plus variés, de nature moins homogène que les autres critères, ce qui rend plus difficile l'apprentissage des caractéristiques. Aussi, il serait nécessaire de faire un apprentissage sur un ensemble plus important d'avis pour améliorer les résultats. La seconde raison réside vraisemblablement dans le fait que quatre personnes ont participé à l'annotation des avis et leur interprétation du critère *Expérience Utilisateur* peut influencer les performances du modèle.

|  | Précision | Rappel | F1 |
|---|---|---|---|
| Évaluation | 0,88 | 0,93 | 0,91 |
| Rapport d'erreur | 0,92 | 0,93 | 0,93 |
| Demande de fonctions | 0,85 | 0,83 | 0,84 |
| Expérience utilisateur | 0,81 | 0,73 | 0,77 |
| **Moyenne pondérée** | 0,88 | 0,89 | 0,89 |

TABLEAU 3 – Résultats de la classification des avis des trois applications

### 3.2 Apprentissage à partir de deux applications

Dans cette expérimentation, nous avons utilisé la même stratégie d'échantillonnage stratifié : 60% des avis de deux applications comme base d'apprentissage, 20% comme validation. Les tests s'appliquent sur les 20% restants et également sur tous les avis de la troisième application. Nous avons fait ceci pour les trois combinaisons possibles d'applications sélectionnées pour l'apprentissage versus le test. Par exemple, une de ces combinaisons a consisté à réaliser l'apprentissage / validation sur 1600 avis de Garmin Connect et 1600 de Huawei Health et le test sur 800 avis de ces applications et les 2000 avis de Samsung Health. Pour chaque combinaison, 10 expériences ont été réalisées. Les résultats moyens des 30 apprentissages ($3 \times 10$) sont présentés dans les Tableaux 4 et 5.

|  | Précision | Rappel | F1 |
|---|---|---|---|
| Évaluation | 0,88 | 0,93 | 0,90 |
| Rapport d'erreur | 0,91 | 0,93 | 0,92 |
| Demande de fonctions | 0,85 | 0,81 | 0,83 |
| Expérience utilisateur | 0,79 | 0,72 | 0,76 |
| **Moyenne pondérée** | 0,87 | 0,88 | 0,88 |

TABLEAU 4 – Résultats de classification des avis pour les 20% restants de deux applications

|  | Précision | Rappel | F1 |
|---|---|---|---|
| Évaluation | 0,88 | 0,92 | 0,90 |
| Rapport d'erreur | 0,85 | 0,92 | 0,88 |
| Demande de fonctions | 0,80 | 0,74 | 0,75 |
| Expérience utilisateur | 0,77 | 0,69 | 0,73 |
| **Moyenne pondérée** | 0,86 | 0,86 | 0,85 |

TABLEAU 5 – Résultats de classification des avis : apprentissage sur deux applications et test sur la troisième application

Les résultats présentés dans le Tableau 4 sont légèrement inférieurs à ceux du Tableau 3. Ils montrent que l'on peut obtenir de bonnes performances de classification avec un jeu de données restreint. Dans le Tableau 5, on peut obser-

ver que la moyenne de la précision et du rappel pour lesquels les modèles n'ont pas été entraînés sont inférieurs de 1-2% par rapport aux modèles ayant servi pour l'entraînement. Cette expérimentation montre que les résultats restent de très bonne qualité en entraînant le modèle sur un sous-ensemble d'applications.

## 4 Discussion et travaux futurs

Dans ce travail, nous avons créé un jeu de données pour la classification multi-labels d'avis rédigés en français d'utilisateurs d'applications de suivi d'activité physique. Nous avons utilisé le modèle CamemBERT pour classifier ces avis et les résultats obtenus montrent de bonnes performances. Les expérimentations d'entraînement et de test sur des applications distinctes montrent qu'il est possible de généraliser le modèle sur des applications de même champ applicatif. Ces résultats nous encouragent à poursuivre nos travaux relatifs à l'ingénierie des exigences dirigée par les données pour le développement de différentes versions d'une application de prévention des effets du vieillissement par le suivi de l'activité physique des seniors pour bien vieillir en santé. Nous envisageons d'affiner la classification des demandes de nouvelles fonctionnalités en permettant d'identifier, par une approche non supervisée, les concepts du domaine de l'application cible, comme par exemple, les vitesses et temps de marche, les déséquilibres ou le suivi du sommeil. Cette classification pourrait être présentée visuellement pour indiquer au concepteur quelles sont les fonctionnalités principalement demandées par les utilisateurs.

L'objectif global sera de (i) regrouper les demandes et propositions par catégorie de besoins, (ii) mettre en correspondance ces besoins avec les modèles présents dans le cahier des charges, afin de pouvoir les prendre en compte plus facilement dans la conception.

## Références



\end{document}